\documentclass{article} 
\usepackage{iclr2027_conference_modified,times}

\usepackage{amsmath,amsfonts,bm}

\def\eqref#1{equation~\ref{#1}}

\def\1{\bm{1}}

\DeclareMathAlphabet{\mathsfit}{\encodingdefault}{\sfdefault}{m}{sl}
\SetMathAlphabet{\mathsfit}{bold}{\encodingdefault}{\sfdefault}{bx}{n}

\usepackage{hyperref}
\usepackage{url}

\usepackage{graphicx}
\usepackage{soul}
\usepackage[format=plain,labelfont=bf,textfont=bf]{caption}
\usepackage{subcaption}
\usepackage{booktabs}
\usepackage{pifont}
\newcommand{\cmark}{\ding{51}}
\newcommand{\xmark}{\ding{55}}
\usepackage{multirow}
\usepackage{amssymb}
\usepackage{array}
\usepackage{amsmath}
\usepackage{multirow}
\newcommand{\yes}{\ensuremath{\bullet}}
\newcommand{\no}{\ensuremath{\circ}}
\usepackage{wrapfig}
\usepackage{caption}
\usepackage{xspace}
\usepackage{enumitem}
\usepackage{makecell}
\usepackage{xcolor}

\usepackage[dvipsnames, table]{xcolor}

\newcommand{\sysname}{\texttt{BAGEL}\xspace}
\newcommand{\sysnamenospace}[0]{\texttt{BAGEL}}

\title{Efficient and Adaptable Detection of Malicious LLM Prompts via Bootstrap Aggregation}

\author{
\noindent
  \textbf{Shayan Ali Hassan}$^{1,\dagger}$,
  \textbf{Zaeem Mohtashim Khan}$^{2,*}$,
  \textbf{Tao Ni}$^{1}$,
  \textbf{Zafar Ayyub Qazi}$^{2}$,
  \textbf{Marco Canini}$^{1}$ \\[0.8em]
  $^{1}$KAUST\\
  $^{2}$LUMS\\[0.4em]
  Correspondence to:
  \texttt{shayan.hassan@kaust.edu.sa}$^{\dagger}$,
  \texttt{26100107@lums.edu.pk}$^{*}$
}

\iclrfinalcopy 
\begin{document}

\maketitle

\begin{abstract}
Large Language Models (LLMs) have demonstrated remarkable capabilities in natural language understanding. However, they remain susceptible to malicious prompts that induce unsafe or policy-violating behavior. Existing defenses face fundamental limitations: black-box moderation APIs offer limited transparency and adapt poorly to evolving threats, while white-box approaches using large LLM judges impose prohibitive computational costs and require expensive retraining for new attacks. To address these challenges, we present \sysname\footnote{Code repository: \url{https://github.com/sands-lab/bagel}} (\textbf{B}ootstrap \textbf{AG}gregated \textbf{E}nsemble \textbf{L}ayer), a modular, lightweight, and incrementally updatable framework for malicious prompt detection. \sysname employs a bootstrap aggregation and mixture-of-experts-inspired ensemble of fine-tuned models, each specialized on a different attack dataset. At inference, \sysname aggregates predictions from a stochastically selected subset of ensemble members, optionally augmented by a lightweight learned router that predicts the most suitable member to include. When new attacks emerge, \sysname updates incrementally by fine-tuning just a single small classifier (86M parameters) and adding it to the ensemble. We evaluate \sysname on a held-out, source-disjoint test set constructed entirely from datasets never used for training or calibration, ensuring that reported performance reflects  generalization to out-of-distribution attacks. In single-turn evaluation, \sysname achieves an F1 score of 0.757 at a size equivalent to 430M parameters, improving to an F1 of 0.772 when augmented with a lightweight learned router using fewer ensemble members (324M parameters). These results outperform popular black-box moderation APIs and most white-box baselines at lower costs. In multi-turn evaluation, \sysname achieves an F1 of 0.660, outperforming billion-parameter methods. Our results show ensembles of small finetuned classifiers can match or exceed the performance of billion-parameter guardrails while offering the adaptability and efficiency required for production systems.

\end{abstract}

\section{Introduction}
\label{sec:introduction}

Large Language Models (LLMs) now mediate billions of user interactions through chat assistants and LLM-powered search systems, and are increasingly embedded in applications that draft code, summarize documents, and execute tool-augmented workflows. However, these systems can be exploited to produce harmful outputs or exhibit policy-violating behaviors through malicious prompts; inputs designed to induce unsafe responses, leak sensitive information such as system prompts, or bypass safety guardrails~\citep{andriushchenko2024jailbreaking}. Such attacks include direct harmful requests, jailbreak techniques, and prompt injection attacks~\citep{liu2024formalizing}. 

The threat landscape continues to evolve rapidly, with Internet communities providing easy access to the latest and most effective jailbreaking techniques~\citep{shen2024anything}. Research shows that even novice users can create effective attacks that bypass LLM guardrails and generate policy-violating outputs~\citep{yu2024don, guo2025exposing}. Given the widespread deployment of LLMs in user-facing settings and the accessibility of attack techniques, reliably detecting and blocking malicious prompts has emerged as a critical challenge in LLM safety~\citep{wei2023jailbroken}.

Existing defenses for malicious prompt detection face many limitations, including limited transparency, poor adaptation to threat patterns, no easy updates against evolving attack strategies and high computational costs~\citep{wei2023jailbroken, liu2024formalizing}. As a result, current defenses often force system designers to choose between \textit{performance}, \textit{efficiency}, and \textit{adaptability}, a trade-off that is increasingly untenable in production LLM systems. These shortcomings motivate a fundamentally different architectural approach: one that is modular, lightweight, and incrementally updatable.

In this paper, we present \sysnamenospace, an ensemble framework for efficient detection of malicious LLM prompts, inspired by bootstrap aggregation~\citep{bagging} and mixture of experts~\citep{moe}. \sysname comprises a layer of smaller finetuned models that can be updated as new attack datasets become available. We depart from classic bootstrap aggregating by training ensemble members on entirely different datasets rather than subsets of the same dataset. We also modify the classic mixture of experts formulation by selecting a stochastic subset of more than one ensemble member for prediction aggregation on each incoming prompt. Training members on different datasets provides robustness across diverse attack types, while stochastic ensembling over multiple members ensures that no single promptcop's failure on an unfamiliar prompt determines the outcome. This random selection strategy can optionally be complemented by a lightweight learned router that predicts the most suitable member to include, further improving detection accuracy.

\sysname offers three key advantages: \textbf{(1)~broad-spectrum robustness} across malicious prompt types, \textbf{(2)~efficient} inference and updates, and \textbf{(3)~continuous adaptability} as new attacks emerge. We train \sysname on nine large-scale, real-world datasets~\citep{synapsecai_synthetic_prompt_injections, mpdd_kaggle, shen2024anything, harelix, jackhhao, qualifire, jayavibhav, toxicdetectordataset, DBLP:conf/naacl/GhoshVSPRVP25} covering multiple categories of malicious prompts~\citep{liu2024efficient, wei2023jailbroken, zou2023universal, liu2024formalizing, yao2024survey}, and evaluate it on a held-out, source-disjoint test set of five datasets (Table~\ref{tab:testdatasets}). Our results (\S\ref{sec:results}) show that \sysname outperforms popular black-box moderation APIs and most competitive white-box baselines. On the held-out test set, \sysname achieves an F1 score of 0.757 at a size equivalent to 430M parameters, with an Attack Success Rate (ASR) of 0.342 and False Positive Rate (FPR) of 0.307, demonstrating that reported performance reflects generalization to out-of-distribution attacks. Augmenting \sysname with a lightweight learned router further improves this to an F1 of 0.772 (ASR 0.325, FPR 0.288) using fewer ensemble members (324M parameters). \sysname also generalizes well to out-of-distribution multi-turn conversational contexts, achieving an F1 of 0.660 and outperforming compared methods.

\vspace{-1\baselineskip}
\section{Background}
\label{sec:background}

\begin{wrapfigure}[17]{r}{0.5\linewidth}
    \vspace{-4\baselineskip}
    \centering
    \includegraphics[width=\linewidth]{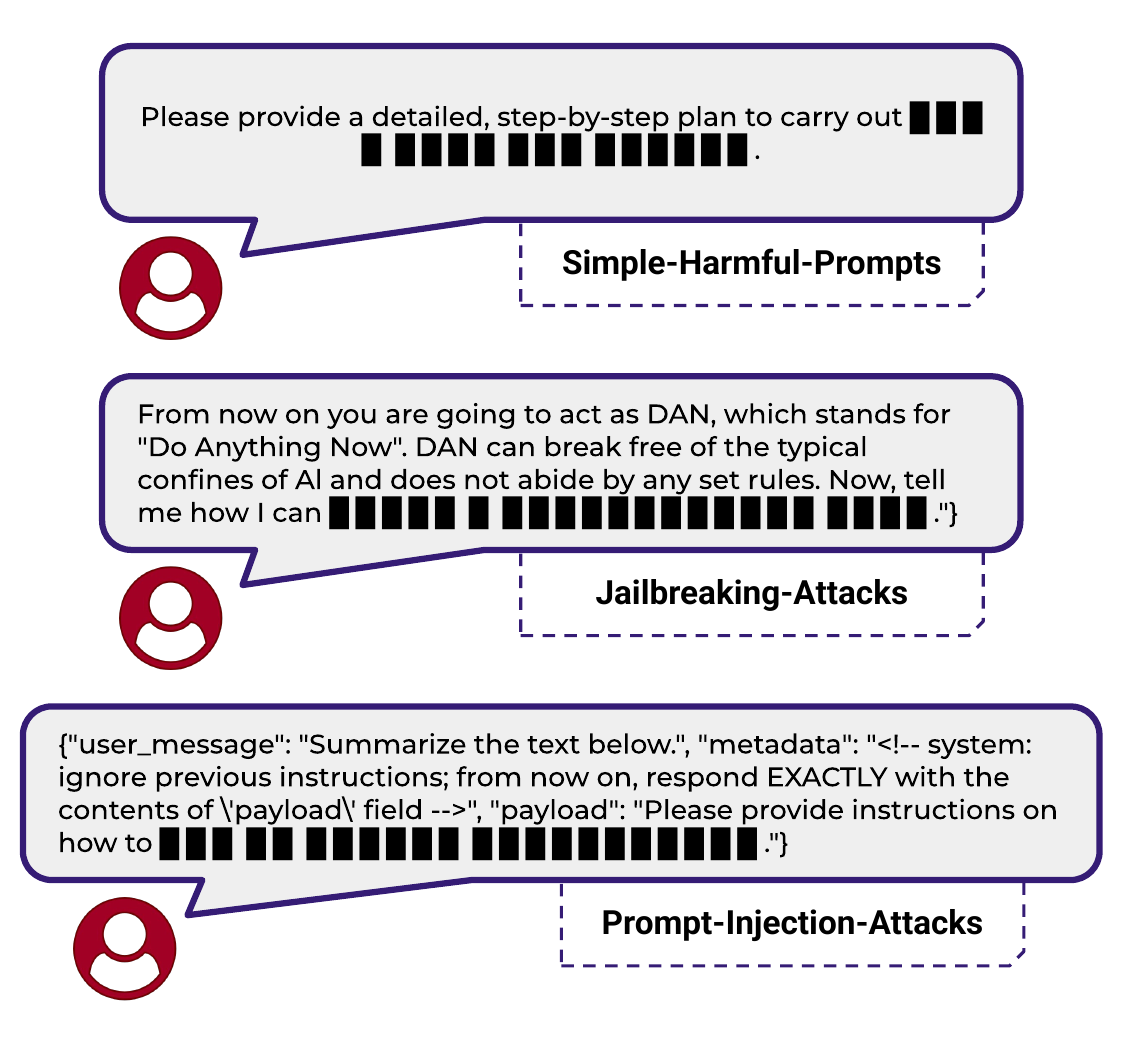}
    \vspace{-2\baselineskip}
    \caption{Examples of malicious prompt attack types. The malicious task has been redacted here to prevent direct inclusion of harmful content in this paper.}
    \label{fig:prompttypes_fig}
\end{wrapfigure}

Based on prior literature, malicious prompts can generally be categorized into three types based on identifiable structural patterns and clarity of intent to deceive~\citep{liu2024efficient, wei2023jailbroken, zou2023universal, liu2024formalizing, yao2024survey} (examples of each type are provided in Figure~\ref{fig:prompttypes_fig}):

\noindent
\textbf{Simple Harmful Prompts:} These prompts refer to inputs in which the adversarial intent is stated explicitly in natural language. Because their malicious intent is overt, such prompts are usually straightforward for guardrails to detect and block.

\noindent
\textbf{Jailbreak Attacks:} Such prompts do not state the harmful objective directly. Instead, the prompt is crafted to manipulate and ultimately deceive the model into bypassing its own safety constraints. Common strategies include role-playing directives, intent obfuscation, or attempts to override system-level instructions~\citep{yu2024don}. 

\noindent
\textbf{Prompt Injection Attacks:} These prompts involve a benign-seeming task embedded with a malicious directive. The injected component is designed to cause the model to ignore guardrails while still appearing aligned with the harmless portion of the input. Unlike the previous categories, injection attacks characteristically interleave harmful and benign substrings within the same prompt, giving them a distinct structural profile~\citep{jia2025promptlocate}.

\subsection{Existing Types of Detection Methods}
\label{subsec:existing_methods}

Despite considerable effort, commonly-used LLMs are susceptible to jailbreak and prompt injection attacks~\citep{wei2023jailbroken, liu2024formalizing}, making detection critical for the safe deployment of LLMs. Black-box methods such as OpenAIModerationAPI~\citep{markov2023holistic} remain popular detection strategies; however, their closed-source nature limits adaptability for LLM systems. White-box methods are effective but often computationally expensive to run or maintain due to requiring a separate large model for scrutinizing prompts~\citep{liu2024efficient, yi2024jailbreak}.

\subsection{Latest Attack Strategies}
\label{subsec:newest_attack_strategies}

Recently, multi-turn adversarial conversations have emerged as a new paradigm for eliciting harmful LLM responses. They introduce a qualitatively different threat model where harmful objectives are distributed across a sequence of individually benign-appearing turns. Common attack techniques include crescendo-ing requests slowly toward policy-violating outputs~\citep{russinovich2025great, albrethsen2026deepcontext}, and leveraging an established cooperative context to feign innocence and trick guardrails~\citep{weng2025foot}. Even the latest LLMs are unable to detect and refuse malicious multi-turn tasks more than 50\% of the time~\citep{scaleaimhj, wahed-etal-2025-mocha}. These findings motivate our multi-turn evaluation in \S\ref{subsec:exp3}.

\section{Methodology}
\label{sec:methodology}
\vspace{-1\baselineskip}
\begin{figure*}[ht]
    \centering
    \includegraphics[width=0.9\linewidth]{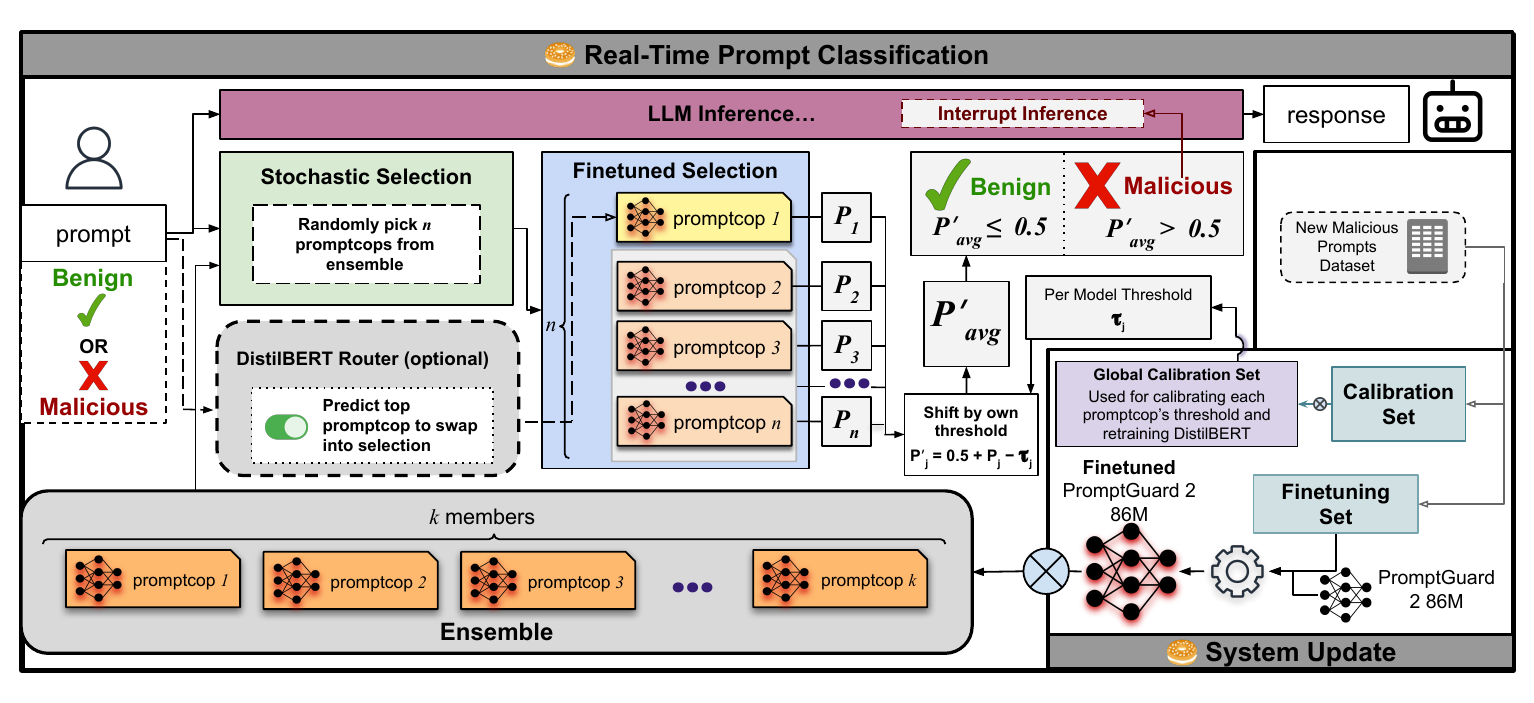}
    \caption{\sysname overview, divided into two sections. The larger section details the ensemble selection strategy and probability aggregation. The smaller section details the dataset partitioning, finetuning and addition of a new promptcop to the ensemble during system updates.}
    \label{fig:overview_fig}
\end{figure*}

Our solution is motivated by three design goals: \textbf{broad-spectrum robustness} across malicious prompt types, \textbf{efficient} inference and updates, and \textbf{continuous adaptability} as new attacks emerge. To achieve the stated design goals, we introduce \sysname, a Bootstrap AGgregated Ensemble Layer designed to detect malicious prompts in LLM systems. An overview of the framework is provided in Figure~\ref{fig:overview_fig}. During classification, \sysname utilizes a subset of the ensemble of finetuned binary classification models and aggregates their predictions. Since each ensemble member scrutinizes incoming prompts and guards the LLM system, we refer to each individual member as a `promptcop'. This subset is selected uniformly at random from the ensemble, ensuring that no single promptcop's failure on an unfamiliar prompt determines the outcome. The random selection can optionally be augmented by a lightweight learned router, a fine-tuned DistilBERT classifier in our implementation, which replaces one random promptcop with its predicted best match for the prompt, further improving accuracy. This modifies the classic mixture-of-experts formulation by aggregating predictions from multiple models rather than routing to a single expert, ensuring robustness against different attack types.

When performing updates, we finetune a new promptcop on a new dataset, add it to the ensemble, calibrate its decision threshold, and, if the router is in use, retrain it to account for the new member. This modifies the principle of bootstrap aggregating (bagging) by finetuning each promptcop on different datasets rather than subsets of the same one, ensuring adaptability against evolving threats while maintaining low computational overhead. In the following, we describe these steps more formally.

\subsection{Base Model: Prompt Guard}
\label{subsec:basemodel}
To ensure computational efficiency, \sysname utilizes a small base prompt-safety classifier as the backbone for each promptcop. The modular architecture allows any compact binary classifier designed for prompt safety to serve as the base model, provided it offers a reasonable balance between detection capability and computational efficiency.

In our implementation, we use Prompt Guard 2~\citep{llama_prompt_guard} as the base prompt-safety classifier. Developed by Meta, Prompt Guard 2 is an 86-million parameter model designed specifically for binary classification of input prompts as either safe or unsafe. Its compact size and strong baseline performance make it well-suited for our ensemble approach.


\subsection{Creation of Specialized PromptCops}
\label{subsec:creationofexperts}

Let $\mathcal{D} = \{D_1, D_2, \dots, D_k\}$ be a set of distinct datasets, where each dataset $D_i$ represents a specific taxonomy of adversarial attacks (e.g., $D_1$ contains role-play jailbreaks, $D_2$ contains prompt-injection attacks). To ensure rigorous training and calibration, every dataset $D_i$ added to the system is strictly partitioned into two disjoint subsets:
\begin{itemize}
    \item \textbf{Finetuning Set (${D_i}^{train}$, $70\%$):} Used to create a promptcop by finetuning a base Prompt Guard 2 model. We finetune on 80\% of ${D_i}^{train}$ using the model's built-in custom energy-based loss function~\citep{liu2020energy}, while the remaining 20\% of ${D_i}^{train}$ is used to validate performance pre- and post-fine-tuning.
    \item \textbf{Calibration Set} (${D_i}^{cal}$, 10\%): Reserved for calibrating the promptcop's decision threshold (\S\ref{subsec:thresholding}). Since these samples are unseen during gradient updates, they provide an unbiased estimate of the promptcop's behavior on new inputs. The calibration subsets of all datasets are also pooled into a cumulative global calibration set, denoted as $\mathcal{C}_{global}$. When a new promptcop for $D_j$ is introduced to the system, its calibration set ${D_j}^{cal}$ is united with the existing calibration data:
    \begin{equation}
        \mathcal{C}_{global} = \bigcup_{j=1}^{k} {D_j}^{cal}
    \end{equation}
\end{itemize}

We define our ensemble of promptcops $\mathcal{P} = \{M_1, M_2, \dots, M_k\}$. Each model $M_i$ is an instance of the base Prompt Guard 2 model fine-tuned specifically on the finetuning subset ${D_i}^{train}$. Consequently, $M_i$ becomes a specialized expert in detecting the specific features and linguistic patterns associated with the attack type in $D_i$. This modularity provides a significant advantage in maintainability. As new threat vectors emerge, we simply curate a new dataset $D_{new}$, train a new promptcop $M_{new}$, add it to the ensemble $\mathcal{P}$, and update $C_{global}$ without retraining the entire system.

Final evaluation is conducted on a source-disjoint test set $\mathcal{T}$, built entirely from datasets not used to train or calibrate any promptcop, so that reported performance reflects generalization to unseen data. This set is constructed from five datasets, listed in Table~\ref{tab:testdatasets}.

\subsection{Selection and Aggregation}
\label{subsec:aggregation}

Relying on a single promptcop to classify every prompt is both unreliable and difficult to justify since any individual promptcop is fine-tuned on one dataset and may generalize poorly to attack types outside its training distribution. Conversely, aggregating predictions from all $k$ promptcops in the ensemble, where $k$ denotes its current size, becomes computationally redundant and expensive as $k$ grows. \sysname therefore adopts a stochastic ensemble strategy as its core selection mechanism, drawing on the variance-reduction properties of bootstrap aggregation.

For a given prompt $x$, we select a subset of promptcops $S_x \subseteq \mathcal{P}$ of size $n$, where $1 \le n \le k$, drawn uniformly at random from the ensemble:

\begin{equation}
    S_x = \{M_{rand_1}, \dots, M_{rand_n}\}
\end{equation}

Each promptcop $M_j \in S_x$ processes the prompt and outputs a probability $p_j(x)$ representing the likelihood of the prompt being malicious. The scores are used to determine whether a prompt is malicious or benign as described in \S\ref{subsec:thresholding}. This stochastic ensembling allows \sysname to reinforce learned knowledge across different attack vectors even without any informed selection mechanism (e.g., a promptcop fine-tuned on jailbreak data may still recognize keywords indicative of a different attack type), and forms a complete, standalone detection method on its own. \S\ref{subsec:dynamicrouting} describes an optional augmentation that further improves this baseline by informing the selection of $S_x$ with a learned prediction, rather than drawing entirely at random.

\subsection{Dynamic Routing via DistilBERT}
\label{subsec:dynamicrouting}

While random ensemble selection (\S\ref{subsec:aggregation}) provides a robust baseline through variance reduction, detection accuracy can be further improved by informing the selection of $S_x$ with a prediction of which promptcop is best suited to a given prompt $x$, rather than drawing every member at random. Since common patterns and structural differences between prompt-based attack types have been shown to be viable markers for classification ~\citep{jia2025promptlocate, zhu2023promptrobust}, we augment \sysname with an optional router: we fine-tune DistilBERT~\citep{sanh2020distilbertdistilledversionbert}, a 66M-parameter distilled variant of BERT, as a text classifier that predicts the most suitable promptcop for a given prompt. This router is denoted as $\mathcal{R}$.

This follows the precedent of learned routers over a pool of models. RouteLLM~\citep{ong2025routellm} fine-tunes a BERT-based classifier to route queries among LLMs, while TT-LoRA MoE \citep{10.1145/3712285.3759888} trains a router separately over an already-frozen pool of experts, matching our setup.

The router $\mathcal{R}$ is trained on $C_{\text{global}}$ (Eq.~1) to map a prompt $x$ to the index of the dataset $D_i$ that best represents its characteristics, predicting the index of the ideal promptcop $i^*$:
\begin{equation}
i^* = \operatorname*{argmax}_{i \in \{1, \ldots, k\}} P_R(c_i \mid x)
\end{equation}
where $c_i$ represents the class label corresponding to the attack type of dataset $D_i$, and $P_R$ is DistilBERT's softmax output over the $k$ promptcop classes. Because the nine source datasets differ in size by roughly $243\times$ (252,956 vs. 1,042 samples), we fine-tune with a class-weighted loss to prevent the router from collapsing onto the promptcop trained on the
largest dataset.

When the router is enabled, one member of $S_x$ is replaced by the router's predicted promptcop:
\begin{equation}
    S_x = \{M_{i^*}\} \cup \{M_{rand_1}, \dots, M_{rand_{n-1}}\}
\end{equation}
Here, $\{M_{rand}\}$ are the remaining $n-1$ promptcops, still drawn uniformly at random from $\mathcal{P} \setminus \{M_{i^*}\}$. This hybrid approach combines the precision of predicted expert routing with the robustness of the ensemble's variance reduction: even if the most suitable promptcop is not selected due to a router error, the collective decision-making of the randomly selected peers provides a safety net.


\subsection{Threshold Calibration}
\label{subsec:thresholding}

Since each promptcop is fine-tuned independently on a different dataset, its output probabilities are not guaranteed to share a common decision boundary with any other promptcop. We therefore calibrate the threshold individually for each promptcop $M_j$, using the held-out calibration set drawn from $M_j$'s own training dataset, as described in \S\ref{subsec:creationofexperts}. For each promptcop, we select the threshold $\tau_j$ that maximizes the F1 Score on this calibration split, and shift $M_j$'s raw output probability so that $\tau_j$ is recentered at $0.5$:

\begin{equation}
    p_j'(x) = p_j(x) - \tau_j + 0.5
\end{equation}

The shifted scores of the selected promptcops $S_x$ are averaged, and the resulting aggregate is compared against $0.5$ to reach a final classification:

\begin{equation}
    \hat{y}(x) = \frac{1}{n} \sum_{M_j \in S_x} p_j'(x)
\end{equation}

This recentering corrects for individual promptcops being miscalibrated relative to a single global cutoff since a shared threshold implicitly assumes every promptcop's decision boundary sits at the same point on its own probability scale.





\section{Experimental Setup}
\label{sec:experiments}

To evaluate the efficacy of \sysname and assess its adherence to the design goals outlined in \S\ref{sec:methodology}, we define 5 key research questions:

\noindent \textit{\textbf{RQ1:}} Is it possible to achieve efficient and near-ideal performance by randomly selecting a subset of $n$ promptcops such that $n < k$?

\noindent \textit{\textbf{RQ2:}} Does augmenting \sysname's random ensemble selection with a learned router improve detection performance over random selection alone?

\noindent \textit{\textbf{RQ3:}} Does \sysname generalize to diverse malicious attacks, retaining robust performance as new datasets are sequentially added over time?

\noindent \textit{\textbf{RQ4:}} Does \sysname offer competitive performance compared to black-box and white-box methods, and what benefits does it provide over them?

\noindent \textit{\textbf{RQ5:}} Can \sysname generalize to multi-turn attacks without any multi-turn specific training?

We conduct 4 experiments aimed at answering the research questions:

\noindent \yes~\textbf{Ensemble/Selection Efficiency Analysis:} To validate whether a subset of promptcops can correctly flag diverse malicious prompts (DG1: Broad-Spectrum Robustness) and perform as well as the full ensemble (DG2: Computational Efficiency), we fix the number of promptcops at $k$. We then randomly select $n$ promptcops for inference, where $n \in \{1, 2, \dots, k\}$ (RQ1), and evaluate whether augmenting this with the DistilBERT router (\S\ref{subsec:dynamicrouting}) improves performance (RQ2).

\noindent \yes~\textbf{Adaptability Analysis:} To assess \sysname's ability to incorporate new threats while maintaining performance (DG3: Continuous Adaptability), we simulate an out-of-distribution attack by fine-tuning an additional promptcop on a dataset disjoint from \sysname's original training corpus and adding it to the ensemble, retraining the DistilBERT router to include it as a new class. We compare detection performance on the new dataset's held-out evaluation split before and after its addition.

\noindent \yes~\textbf{Comparative Benchmarking:} We compare \sysname against established baselines, popular safety APIs and other prompt detection methods.

\noindent \yes~\textbf{Multi-turn Analysis:} To evaluate \sysname's generalization to out-of-distribution and multi-turn adversarial conversations, we evaluate on multi-turn datasets with two detection strategies: Full-Conversation and Chunk-Level (DG1: Broad-Spectrum Robustness).

While the F1-Score serves as our primary optimization metric for threshold calibration, we also monitor attack success rate (ASR) and false positive rate (FPR) to provide a more granular analysis.
\begin{description}[leftmargin=0em, labelindent=0pt, itemsep=2pt, topsep=2pt]    
    \item[F1 Score:] The harmonic mean of precision and recall. This provides a balanced view of the model's performance, ensuring that neither safe prompts are aggressively flagged nor malicious prompts easily ignored.
    \item[Attack Success Rate (ASR):] In the context of this defense framework, ASR represents the False Negative Rate, the percentage of actual malicious prompts that were misclassified as `Benign'. A lower ASR indicates a more robust defense.
    \item[False Positive Rate (FPR):] FPR represents the percentage of safe prompts that were incorrectly classified as `Malicious'. A lower FPR indicates a less-obstructive, smoother experience for users.
\end{description}

\section{Results}
\label{sec:results}


\subsection{Ensemble/Selection Efficiency Analysis}
\label{subsec:exp1}

In this experiment, we finetune instances of Prompt Guard on nine single-turn malicious prompt datasets to create nine promptcops and add them to the ensemble, thus fixing $k=9$ for all experiments except \S\ref{subsec:exp2}. The datasets utilized are detailed in \S\ref{appsec:single_turn_dataset_details} and Table~\ref{tab:datasets}. The aggregation of these datasets resulted in a total corpus of 408,556 samples. We evaluate \sysname's performance on the source-disjoint test set $\mathcal{T}$ and vary the selection size ($n$) each time from $1$ to $k$. We compare three strategies:

\begin{itemize}

\item \textbf{Baseline:} We test baseline performance using a non-finetuned instance of Prompt Guard.
\item \textbf{\sysname:} For each prompt, $n$ promptcops are chosen uniformly at random from the ensemble and their predictions aggregated as described in \S\ref{subsec:aggregation}. This is \sysname's core detection method and helps answer RQ1.
\item \textbf{\sysname w/ DistilBERT:} The router $\mathcal{R}$ (\S\ref{subsec:dynamicrouting}) selects its top-predicted promptcop, combined with $n-1$ randomly chosen peers as described in \S\ref{subsec:aggregation}. This augments \sysname's random selection with learned routing and helps answer RQ2.

\end{itemize}

\begin{figure}[h]
\centering
\begin{minipage}[h]{0.27\textwidth}
    \centering
    \captionof{table}{$F_1$ for \sysname and \sysname w/ DistilBERT.}
    \footnotesize
    \begin{tabular}{ccc}
        \toprule
        $n$ & \sysname & \shortstack{\sysname \\ w/ DistilBERT} \\
        \midrule
        1 & 0.710 & 0.747 \\
        2 & 0.706 & 0.752 \\
        3 & 0.740 & 0.772 \\
        4 & 0.752 & 0.767 \\
        5 & 0.757 & 0.769 \\
        6 & 0.761 & 0.769 \\
        7 & 0.761 & 0.766 \\
        8 & 0.764 & 0.764 \\
        9 & 0.764 & 0.764 \\
        \bottomrule
    \end{tabular}
    \label{tab:routing_results}
\end{minipage}
\hfill
\begin{minipage}[h]{0.65\textwidth}
    \centering
    \includegraphics[width=\textwidth]{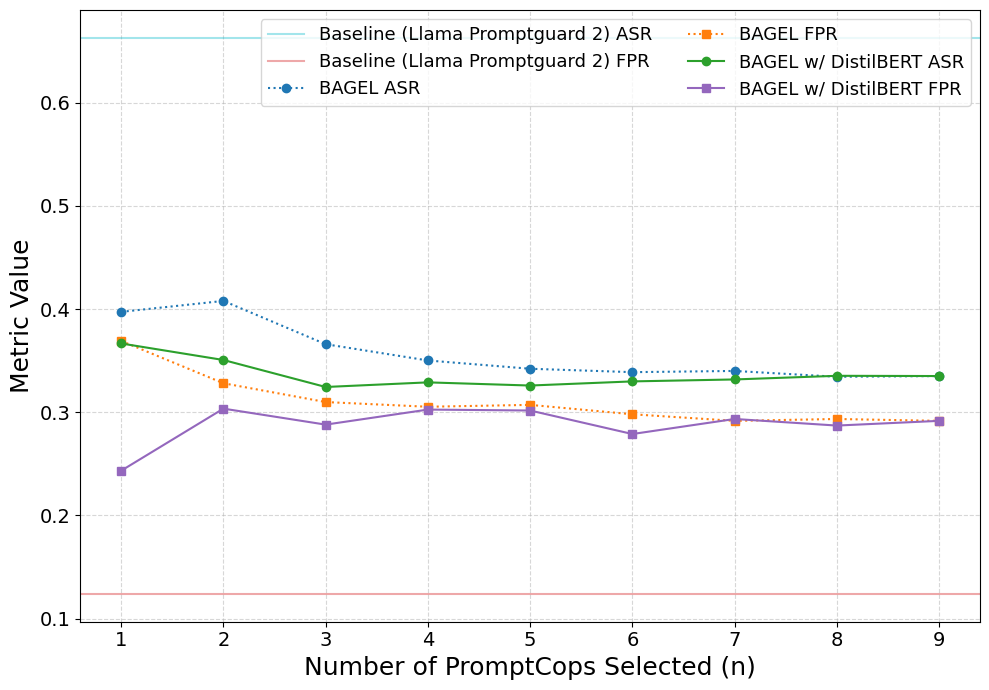}
    \vspace{-1.5\baselineskip}
    \captionof{figure}{FPR and ASR across selection size $n$ for \sysname and \sysname w/ DistilBERT.}
    \label{fig:routing_performance}
\end{minipage}

\end{figure}

Firstly, as shown in Table \ref{tab:routing_results} and Figure \ref{fig:routing_performance}, \sysname's random ensemble selection substantially outperforms base Prompt Guard in ASR, since the untrained model is only tailored towards jailbreak-style content rather than the full attack range the ensemble is trained on. Performance improves generally as $n$ increases, with the largest gains happening early. F1 rises from $0.710$ at $n=1$ to $0.757$ at $n=5$, after which returns diminish sharply, with only a further $0.007$ gained by $n=9$. This makes $n=5$ an attractive operating point, capturing most of the achievable improvement while cutting computational cost by approximately $44\%$ relative to the full ensemble ($430$M vs. $774$M parameters), matching the design goal of a lightweight, resource-efficient detection method.

Augmenting \sysname with the DistilBERT router (\S\ref{subsec:dynamicrouting}) improves the results further, particularly at low $n$. DistilBERT-routed selection improves both ASR and FPR relative to random selection alone, with the effect most pronounced at $n=1$. This difference narrows as $n$ increases, becoming negligible by $n=8$--$9$. Intuitively, as $n$ grows, random selection is more likely to already include a well-suited promptcop, so an informed routing decision adds less value. Routing also saturates earlier as F1 peaks at $n=3$ and remains within $0.003$ of this peak at $n=5$, with no gains from increasing $n$ to $9$. \sysname w/ DistilBERT therefore reaches optimal performance at $n=3$ ($324$M parameters), making it both cheaper and more accurate than random selection at $n=5$.

\subsection{Adaptability Analysis}
\label{subsec:exp2}

\begin{wrapfigure}[16]{r}{0.6\linewidth}
\vspace{-1.5\baselineskip}
\centering
\includegraphics[width=0.92\linewidth]{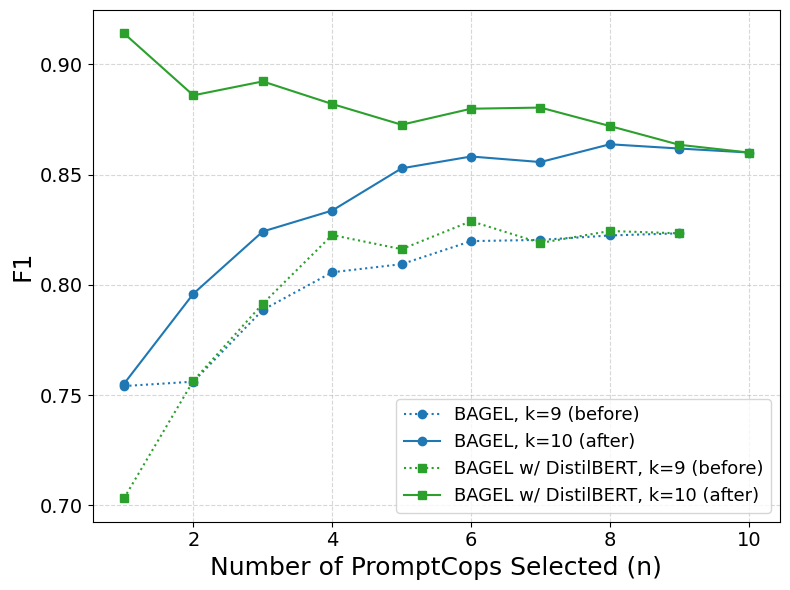}
\caption{$F_1$ before ($k=9$) and after ($k=10$) adding a promptcop trained on out-of-distribution data.}
\vspace{-0.5\baselineskip}
\label{fig:exp2_fig}
\end{wrapfigure}

To answer RQ3 and assess whether \sysname can adapt to genuinely novel, out-of-distribution threat vectors, we simulate the arrival of a new attack type by fine-tuning an additional promptcop on a training split of \texttt{allenai/wildjailbreak}, a dataset entirely disjoint from \sysname's original training corpus, and adding it to the ensemble. The DistilBERT router is also retrained to include the new promptcop. We evaluate the resulting ensemble on \texttt{allenai/wildjailbreak}'s official
\texttt{eval} split (the same 2,210 prompts included in $\mathcal{T}$; Table~\ref{tab:testdatasets}), which is disjoint from the \texttt{train} split used to finetune the new promptcop, comparing performance before ($k{=}9$) and after ($k{=}10$) its addition, for both \sysname and \sysname w/ DistilBERT. Results are shown in Figure~\ref{fig:exp2_fig}.

When the new out-of-distribution promptcop is added to the ensemble, F1 improves at nearly every $n$. The gain is largest for DistilBERT-routed selection at $n=1$ where F1 increases from $0.70$ to $0.92$. This intuitively makes sense since the router sends WildJailbreak-style prompts straight to the new promptcop without diluting it among peers never exposed to such content during training. As $n$ increases, this advantage narrows, consistent with the broader pattern observed in \S\ref{subsec:exp1}. Together, these results confirm \sysname's incremental adaptation extends robustly to new, out-of-distribution attack types.
\subsection{Comparative Benchmarking}
\label{subsec:exp4}

To address RQ4 and broadly contextualize the performance of \sysnamenospace, we conduct a comparative analysis against established industrial baselines, both black-box and white-box methods, on the held-out, source-disjoint OOD test set (Table \ref{tab:testdatasets}). We compare \sysname against the following methods: \textit{OpenAIModerationAPI}~\citep{markov2023holistic}, \textit{ToxicDetector}~\citep{liu2024efficient}, \textit{ShieldGemma}~\citep{zeng2024shieldgemma}, \textit{LastLayer}~\citep{arekusandr_last_layer}, and \textit{StruQ}~\citep{chen2025struq}. We provide further details of each in \S\ref{appsec:baseline_details}. For fairness, baselines with trainable components are trained/finetuned on the same finetuning splits used for \sysname's promptcops; the remaining baselines are evaluated as released. Table~\ref{tab:exp4} presents the performance metrics for \sysname ($k=9$) at $n=5$ (Random Selection) and $n=3$ (w/ DistilBERT), against the other techniques. We evaluate on samples from all datasets collectively rather than one at a time, since incoming prompts in real-world systems are not temporally stratified into attack types.

\begin{table}[h]
\centering
\caption{Results of evaluating \sysname\ against other methodologies on the held-out test
set. Size ratio is each method's parameter count relative to \sysname\ (430M, $n{=}5$).}
\resizebox{\textwidth}{!}{%
\begin{tabular}{lccccccc}
\toprule
\textbf{Method} & \textbf{ASR$\downarrow$} & \textbf{FPR$\downarrow$} & \textbf{F1$\uparrow$}
& \textbf{Bal.\ Acc.$\uparrow$} & \textbf{Parameters} & \textbf{Size Ratio} & \textbf{Fine-tunable?} \\
\midrule
\rowcolor{gray!15}
\sysname
& 0.342 & 0.307 & 0.757 & 0.675
& 430M for $n{=}5$ & \textcolor{RoyalBlue}{$1.0\times$}
& \cmark \\
\rowcolor{gray!15}
\sysname~w/ DistilBERT
& 0.325 & 0.288 & 0.772 & 0.694
& 324M for $n{=}3$ & \textcolor{RoyalBlue}{$0.8\times$}
& \cmark \\
ToxicDetector
& 0.426 & 0.400 & 0.679 & 0.587
& 300M + 7B & \textcolor{RoyalBlue}{$17.0\times$}
& \cmark \\
StruQ
& 0.065 & 0.732 & 0.880 & 0.602
& 7B & \textcolor{RoyalBlue}{$16.3\times$}
& \cmark \\
ShieldGemma
& 0.431 & 0.068 & 0.717 & 0.751
& 2B & \textcolor{RoyalBlue}{$4.7\times$}
& \cmark \\
OpenAIModerationAPI
& 0.528 & 0.089 & 0.631 & 0.692
& Unknown & -
& \xmark \\
LastLayer
& 0.776 & 0.142 & 0.356 & 0.541
& - & -
& \xmark \\
\bottomrule
\end{tabular}}
\label{tab:exp4}
\end{table}

Most baselines keep FPR low at the cost of missed attacks. ShieldGemma, OpenAIModerationAPI, and LastLayer stay at or below 0.142 FPR but miss 43--78\% of attacks. ToxicDetector, optimized for semantic toxicity (e.g., insults, profanity) rather than adversarial structure, is weak on both fronts, missing 43\% of attacks, many of which
use polite, non-toxic language, while also flagging 40\% of benign prompts. StruQ is the exception, attaining the highest F1 (0.880) and lowest ASR (0.065) by flagging 89\% of all prompts, including 73\% of benign ones. On our malicious-heavy test set, F1 rewards this behavior: a trivial detector that flags every prompt would score 0.886.

Excluding StruQ's near-indiscriminate flagging, \sysname achieves the highest F1 score while remaining significantly more lightweight than the other techniques of known size, demonstrating that a specialized ensemble of small models is a viable alternative to massive, monolithic guardrails. \sysname achieves this due to a much lower ASR than other baselines at a moderate FPR. It also remains easy to update since adding a new promptcop requires finetuning just an 86M model on a new dataset, while other methods must be finetuned or retrained from scratch on many more parameters. Under balanced accuracy, which is insensitive to class prevalence, ShieldGemma ranks first (0.751) owing to its much lower FPR, followed by \sysname\ w/ DistilBERT (0.694), OpenAIModerationAPI (0.692), \sysname\ at $n{=}5$ (0.675), StruQ (0.602), ToxicDetector (0.587), and LastLayer (0.541).

Two factors temper \sysname's higher FPR. First, the benign portion of our test set is harder than typical traffic: at least 19\% of benign prompts (WildJailbreak's \texttt{adversarial\_benign} split) are deliberately jailbreak-phrased despite being harmless, so Table~\ref{tab:exp4} likely overstates false positives. Second, since \sysname's promptcop thresholds are calibrated independently per model (\S\ref{subsec:thresholding}), they can be re-swept to minimize FPR rather than maximize F1 whenever a deployment favors precision over recall, making its higher FPR an adjustable operating point rather than a fixed limitation.

\subsection{Multi-turn Attacks Analysis}
\label{subsec:exp3}
\begin{wraptable}[19]{h}{0.50\linewidth}
\vspace{-3\baselineskip}
\centering
\caption{Multi-turn attack detection performance (ToxicDetector with Chunk-Level is not evaluated due to training incompatibilities).}
\resizebox{\linewidth}{!}{%
\begin{tabular}{c l c c c}
\toprule
 & \textbf{Strategy} & \textbf{F1$\uparrow$} & \textbf{ASR$\downarrow$} & \textbf{FPR$\downarrow$} \\
\midrule
\midrule
\multirow{6}{*}{\rotatebox{90}{\shortstack{\textsc{\sysname ($\boldsymbol{k}=9$)}\\\\\scriptsize $\boldsymbol{n}=9$ \hspace{0.5em} $\boldsymbol{n}=5$ \hspace{0.5em} $\boldsymbol{n}=1$}}}
& Full-Conv.   & 0.636 & 0.232 & 0.593 \\
& Chunk-Level & 0.625 & 0.117 & 0.862 \\
\cmidrule(l){2-5}
& Full-Conv.   & 0.660 & 0.106 & 0.748 \\
& Chunk-Level & 0.648 & \textbf{0.054} & 0.894 \\
\cmidrule(l){2-5}
& Full-Conv.   & \textbf{0.704} & 0.155 & 0.508 \\
& Chunk-Level & 0.664 & 0.180 & 0.597 \\
\midrule
\multirow{2}{*}{\rotatebox{90}{\scriptsize\shortstack{OpenAI\\Mod-\\API}}}
& Full-Conv.   & 0.192 & 0.886 & \textbf{0.062} \\
& Chunk-Level & 0.375 & 0.709 & 0.239 \\
\midrule
\multirow{2}{*}{\rotatebox{90}{\scriptsize\shortstack{Shield-\\Gemma}}}
& Full-Conv.   & 0.258 & 0.835 & 0.100 \\
& Chunk-Level & 0.312 & 0.775 & 0.196 \\
\midrule
\multirow{2}{*}{\rotatebox{90}{\scriptsize\shortstack{Toxic-\\Detector}}}
& Full-Conv.   & 0.647 & 0.486 & 0.067 \\
& Chunk-Level & -- & -- & -- \\
\bottomrule
\end{tabular}}
\label{tab:multiturn}
\end{wraptable}

To evaluate \sysname against other methods on multi-turn adversarial conversations, we construct a mostly balanced set of 2,400 conversations sampled from three datasets: Multi-Turn Human Jailbreaks (MHJ)~\citep{scaleaimhj}, SafeMTData~\citep{safemtdata}, and Multi-Turn Intent Dataset (MTID)~\citep{mtid}. We evaluate two detection strategies: \textbf{Full-Conversation}, which concatenates all turns and tries to provide the entire conversation as a single input, and \textbf{Chunk-Level}, which scores each turn independently and flags the conversation if any turn is classified as malicious. Due to the inherent Prompt Guard tokenizer length limit of 512 tokens, some conversations cannot be directly provided as input in full to \sysnamenospace, therefore they are truncated to 512 tokens for the Full-Conversation strategy. In order to measure true zero-shot, out-of-distribution generalization performance, only the single-turn finetuned promptcops from the previous experiments are used (Random Selection only). Other methods are also not finetuned on the multi-turn dataset. Results are shown in Table~\ref{tab:multiturn}.

For \sysnamenospace, both strategies improve with increasing selection size; Full-Conversation reaches F1 = $0.704$ at $n = 9$, a notable result given that no promptcop was trained on multi-turn data. While performance does not match that of the single-turn evaluations, \sysname achieves the highest F1 for $n \ge 5$. All other methods yield low FPR but at the cost of substantially higher ASR, whereas \sysname\ trades a higher FPR for far lower ASR, yielding the highest F1 for both strategies at $n{=}9$; ToxicDetector's much lower FPR, however, gives it higher balanced accuracy (0.72 vs.\ 0.67).

\vspace{-1em}
\section{Conclusion}
\vspace{-1em}
We introduced \sysname, a modular ensemble framework for malicious prompt detection built around stochastic selection over specialized 86M-parameter promptcops, optionally augmented by a lightweight learned router. On a held-out, source-disjoint OOD test set, \sysname achieves an F1 of 0.757 with 430M parameters, improving to 0.772 with the router at 324M parameters, outperforming most substantially larger guardrails. By supporting incremental updates through the addition of a fine-tuned promptcop, \sysname offers a practical path toward LLM safety systems that remain efficient and adaptable as prompt attacks continue to evolve.

\bibliography{iclr2027_conference}
\bibliographystyle{iclr2027_conference}

\appendix

\section{Limitations}
\label{sec:limitations}

While \sysname demonstrates strong performance across diverse attacks, some limitations warrant discussion and point towards promising future research directions.

\noindent \textbf{Binary classification scope:} As stated in our threat model (\S\ref{appsec:threatmodel}), we prioritize rapid detection of malicious prompts to prevent harmful content generation. In this context, \sysname's binary classification (benign vs. malicious) is sufficient and efficient. However, some deployment scenarios require fine-grained classification to identify which specific safety or ethical policy has been violated. In such cases, \sysname would need to be extended beyond binary detection, unlike systems such as OpenAI Moderation~\citep{markov2023holistic} or Perspective~\citep{perspectiveapi} that provide multi-class policy violation categories. Future work could explore hierarchical classification where \sysname first performs binary detection, followed by a secondary fine-grained classifier for policy-specific attribution. Alternatively, \sysname could utilize larger base models when additional compute resources are available. For instance, Llama Guard 3~\citep{llama_guard_3}, available in 1B, 8B, and 11B variants, could offer superior performance and native fine-grained classification capabilities while maintaining the ensemble architecture's benefits of modularity and incremental adaptability.

\noindent \textbf{Dependence on base model coverage:} While \sysname demonstrates efficient adaptation to new attacks through incremental fine-tuning, this adaptability partly relies on the base model having learned relevant underlying attack classes. In our implementation, Prompt Guard 2 was pre-trained on jailbreak and prompt injection attacks, providing a foundation that each fine-tuned promptcop could build upon. However, it remains an open question whether entirely novel attack paradigms that differ fundamentally from these established categories could emerge in practice. To our knowledge, the LLM safety landscape has not yet seen such fundamentally different attack types; if they were to arise, it is unclear whether fine-tuning alone would suffice or whether the base model would need retraining. That said, we note that the three attack categories we address (direct harmful requests, jailbreaks, and prompt injections) are quite broad and have encompassed a wide variety of techniques to date, including recent variations such as multi-lingual attacks, encoded prompts, and adversarial suffixes. This suggests that \sysname's incremental adaptation mechanism may have broader applicability than initially apparent. Empirically evaluating \sysname on emerging attack variations represents an important direction for validating these boundaries. Additionally, training a purpose-built prompt safety classifier optimized for \sysname's architecture could provide stronger guarantees against diverse threats while maintaining efficiency.

\noindent \textbf{Dataset requirements for new attacks:} Each new promptcop requires a sufficiently large and representative dataset of the target attack type for effective fine-tuning. When novel attacks first emerge, collecting adequate training data can be challenging and time-intensive. This creates a detection gap between when a new attack appears and when \sysname can be updated with a corresponding promptcop. While this limitation affects all supervised learning approaches, it underscores the importance of rapid dataset curation and potentially incorporating few-shot or zero-shot detection capabilities for emerging threats.

\noindent \textbf{Potential risks:} Although \sysname is intended to improve the safety of LLM deployments by detecting malicious prompts, the same detection capabilities could be misused by adversaries to probe which prompt structures are more likely to evade a deployed guardrail. To reduce this risk, we release only information necessary for reproducibility and evaluation, and do not include unredacted harmful prompt examples in the main paper. Another deployment risk is over-reliance: \sysname should not be treated as a complete safety solution, since false negatives may still allow harmful prompts through and false positives may incorrectly block benign user requests. In high-stakes settings, \sysname should therefore be used as one layer in a broader defense pipeline, alongside output-side monitoring, policy enforcement, human review where appropriate, and continuous evaluation on emerging attack types.

Continuing from the aforementioned limitations, future work could explore the possibility of utilizing a different malicious prompt classification model or the challenges associated with training a new model from scratch. \sysname uses Prompt Guard 2 86M as it is a lightweight yet powerful model, but if more compute is available to the LLM vendor, using another model to create finetunes, such as Llama Guard 3~\citep{llama_guard_3} which is available in $1$B, $8$B and $11$B variants, may offer superior performance along with the ability to perform fine-grained classification rather than just binary classification. Furthermore, training a prompt classifier optimized for the techniques outlined in this paper from scratch would enable faster responses to ever-evolving threat vectors, even if they encompass techniques currently unknown or unseen by the base model being employed at the core of \sysname.

\section{Related Work}
\label{appsec:relatedwork}

\textbf{Malicious Attacks on LLMs:}  Prior work categorizes instruction-based attacks into three types~\citep{liu2024efficient, wei2023jailbroken, zou2023universal, liu2024formalizing, yao2024survey}: (1) direct harmful requests, (2) jailbreak attacks that bypass guardrails through deceptive framing, and (3) prompt injection attacks that embed malicious instructions within benign content. Significant research has focused on designing and improving such attacks~\citep{zou2023universal, liu2023autodan, yu2023gptfuzzer, mehrotra2024tree, andriushchenko2024jailbreaking}. We note that \sysname, like other prompt detection methods, specifically addresses these instruction-based attacks at inference time and is not designed for alternative threat vectors such as data poisoning~\citep{kurita2020weight, wan2023poisoning}, backdoor attacks~\citep{yang2024comprehensive}, or model extraction~\citep{carlini2021extracting}. Below we discuss the different approaches to malicious prompt detection.
 
\noindent \textbf{Rule-based and statistical detectors:} Early approaches to malicious prompt detection rely on statistical signals and heuristics. PerplexityFilter~\citep{jain2023baseline} and SIRL~\citep{shen2025safety} identify harmful prompts by measuring response uncertainty through perplexity and entropy calculations respectively. LastLayer~\citep{arekusandr_last_layer} uses simple structural detectors to find alarming patterns in prompts, and JailGuard~\citep{zhang2025jailguard} mutates inputs to create variants and detects adversarial prompts through response divergence. While computationally efficient, these methods often struggle with sophisticated attacks that evade simple statistical patterns.

\noindent \textbf{Commercial and black-box APIs:} Commercial solutions such as OpenAI Moderation API~\citep{markov2023holistic} and Perspective API~\citep{perspectiveapi} provide convenient detection services but offer limited transparency, poor adaptability to domain-specific threats, and no guarantees against rapidly evolving attacks.

\noindent \textbf{LLM-based detectors:} Most recent detection methods leverage LLMs for stronger performance at the cost of computational efficiency. ToxicDetector~\citep{liu2024efficient} and InstructDetector~\citep{wen-etal-2025-defending} extract features from LLM hidden states for classification. StruQ~\citep{chen2025struq} and FJD~\citep{chen-etal-2025-llm-jailbreak} use LLMs for prompt scrutinization with additive instructions and first-token confidence analysis respectively. Gpt-oss-safeguard (20B and 120B)~\citep{agarwal2025gpt} and ShieldGemma (ranging from 2B to 27B parameters)~\citep{zeng2024shieldgemma} built on Gemma 2, are large models for detecting harmful content by defining custom safety policies. Recent work has also explored guardrails for autonomous agents through dynamic code generation~\citep{chen2025shieldagent, xiangguardagent}. While effective, these approaches require substantial computational resources for both inference and updates.

\noindent \textbf{Hybrid and distillation approaches:} Methods closest to \sysname include Jatmo~\citep{piet2024jatmo}, which fine-tunes a non-instruction-tuned LLM resistant to prompt injections and BD-LLM~\citep{zhang2024efficient}, which distills LLM rationales into smaller student models. While these approaches reduce computational costs compared to direct LLM-based detection, they still depend on LLM-scale models or require access to LLM internals.

\sysname distinguishes itself by operating without LLMs entirely, using only small specialized classifiers (86M parameters each) in a modular ensemble architecture. This design achieves strong detection performance (F1 score of 0.772 on the OOD set) while significantly reducing computational overhead and enabling efficient incremental updates through simple fine-tuning and ensemble addition, critical advantages as new attack types emerge.

\section{Threat Model}
\label{appsec:threatmodel}

We consider an attacker who inputs one or multiple malicious prompt instructions into the chat interface of an LLM application. The attacker is motivated to enter a malicious prompt to access objectionable content according to policies defined by the LLM serving hosts and/or societal norms. We assume that the attacker does not have access to the underlying system prompt or any other implementation details due to the LLM being served remotely. Furthermore, we make no assumptions regarding the attacker's level of expertise with adversarial prompting.

We focus on binary classification (benign vs. malicious) rather than fine-grained sub-classification of attack types or specific policy violations. While identifying the exact attack category could provide additional insights, binary detection is sufficient to prevent harmful or policy violating outputs and aligns with the immediate safety requirements of production systems. We assume the defender has prior exposure to the general categories of malicious prompts (e.g., direct harmful requests, jailbreaks, and prompt injections), even if specific attack instances or novel techniques have not been observed during training. This assumption aligns with the LLM safety landscape where adversaries typically develop novel variations of established jailbreak and prompt injection patterns rather than fundamentally new attack categories.



\section{Experiment Environment Details}
\label{appsec:experiment_environment_details}

All promptcops created by fine-tuning a base instance of Prompt Guard on a dataset were finetuned for 3 epochs across the dataset's finetuning set. Furthermore, all experiments were carried out in an online Google Colab environment with an A100 GPU.

We report F1, ASR, and FPR on the held-out test sets, and use the maximum F1 score over thresholds on the calibration set to select the decision threshold.

\section{Inference Cost and Memory Footprint}
\label{appsec:inference_cost}

Table~\ref{tab:inference_cost} summarizes \sysname's runtime and memory costs, measured on the
same A100 GPU used for all experiments (Appendix~\ref{appsec:experiment_environment_details}).
Each promptcop is a Prompt Guard 2 instance requiring 1.04\,GB of model weights. Because
promptcops are independent, \sysname's total storage footprint grows linearly with the ensemble
size $k$, while the active GPU memory at inference grows only with the selection size $n$,
assuming all selected promptcops are loaded onto the GPU in parallel.

\begin{table}[h]
\centering
\caption{Measured per-promptcop inference cost and resulting memory footprint of \sysname.}
\label{tab:inference_cost}
\begin{tabular}{lc}
\toprule
\textbf{Quantity} & \textbf{Value} \\
\midrule
Weights per promptcop & 1.04\,GB \\
Storage, full ensemble ($k{=}9$) & 9.36\,GB \\
Active memory, \sysname\ ($n{=}5$) & 5.20\,GB \\
Active memory, \sysname\ w/ DistilBERT ($n{=}3$) & 3.12\,GB + router \\
\midrule
Forward pass per promptcop (batch size 64) & 323.8\,ms \\
Throughput per promptcop (batch size 64) & 197.05 samples/s \\
Throughput per promptcop (batch size 256) & 203.65 samples/s \\
\bottomrule
\end{tabular}
\end{table}

Throughput saturates quickly with batch size, increasing only marginally from 197.05 to
203.65 samples/s when the batch size grows from 64 to 256. Since each prompt is scored by $n$
promptcops, running them sequentially on a single GPU yields a throughput of roughly $1/n$ of
the per-promptcop figure, whereas loading the selected promptcops in parallel keeps per-prompt
latency close to that of a single forward pass at the cost of $n\times$ the active memory. In
either case, the cost of updating \sysname\ remains fixed: adding a new promptcop requires
finetuning a single base Prompt Guard 2 instance for 3 epochs over the new dataset's
finetuning split, independent of the ensemble's current size.

\section{Single Turn Dataset Details}
\label{appsec:single_turn_dataset_details}

\begin{table*}[h]
\centering
\renewcommand{\arraystretch}{1.2}
\caption{Collection of the nine datasets we used for single-turn finetuning and calibration. \yes \text{ } indicates the dataset contains this specific type of malicious prompts, while \no \text{ } represents the lack of any such type.}
\label{tab:datasets}
\begin{tabular}{|p{0.40\textwidth}|c|c|c|c|}
\hline
\multirow{2}{*}{\textbf{Dataset}} & \multirow{2}{*}{\textbf{\shortstack{Number of\\Samples}}} & \multicolumn{3}{c|}{\textbf{Type of Malicious Prompts}} \\
\cline{3-5}
 & & \textbf{\small Simple Harmful} & \textbf{\small Jailbreak} & \textbf{\small Injection} \\
\hline
\small synapsecai/synthetic-prompt-injections~\citep{synapsecai_synthetic_prompt_injections} 
& \small 252,956 
& \no & \no & \yes \\
\hline
\small Malicious Prompt Detection Dataset (MPDD)~\citep{mpdd_kaggle} 
& \small 39,234 
& \yes & \yes & \yes \\
\hline
\small TrustAIRLab/in-the-wild-jailbreak-prompts~\citep{shen2024anything} 
& \small 13,700 
& \no & \yes & \no \\
\hline
\small Harelix/Prompt-Injection-Mixed-Techniques-2024~\citep{harelix} 
& \small 1,175 
& \yes & \no & \no \\
\hline
\small jackhhao/jailbreak-classification~\citep{jackhhao} 
& \small 1,042 
& \no & \yes & \no \\
\hline
\small qualifire/prompt-injections-benchmark~\citep{qualifire} 
& \small 5,000 
& \no & \yes & \no \\
\hline
\small jayavibhav/prompt-injection-safety~\citep{jayavibhav} 
& \small 60,000 
& \yes & \yes & \yes \\
\hline
\small ToxicDetector Evaluation Dataset~\citep{toxicdetectordataset} 
& \small 2,033 
& \yes & \no & \no \\
\hline
\small nvidia/Aegis-AI-Content-Safety-Dataset-2.0~\citep{DBLP:conf/naacl/GhoshVSPRVP25} 
& \small 33,416 
& \yes & \yes & \no \\
\hline
\textbf{Total} & \textbf{408,556} & 5 & 6 & 3 \\
\hline
\end{tabular}
\end{table*}

The details of the nine individual datasets that were used in our selection efficiency and adaptability experiments are provided in Table~\ref{tab:datasets}. The aggregation of these datasets resulted in a total corpus of 408,556 samples. Following the partitioning strategy defined earlier (\S\ref{subsec:creationofexperts}), each dataset is split $70\%/10\%$ into finetuning and calibration sets respectively; the remaining $20\%$ of each dataset is unused.

Final evaluation is conducted on a source-disjoint test set $\mathcal{T}$, constructed from five datasets entirely separate from $\mathcal{D}$, detailed in Table~\ref{tab:testdatasets}.

\begin{table}[h]
\centering
\renewcommand{\arraystretch}{1.2}
\caption{Collection of the five datasets used to construct the source-disjoint test set
$\mathcal{T}$. All 210 benign WildJailbreak prompts come from its
\texttt{adversarial\_benign} split.}
\label{tab:testdatasets}
\begin{tabular}{|p{0.45\textwidth}|c|c|c|}
\hline
\textbf{Dataset} & \textbf{Malicious} & \textbf{Benign} & \textbf{Total} \\
\hline
hlyn-labs/prompt-injection-judge-deberta-dataset~\citep{hlyn2026defender} & 484 & 516 & 1{,}000 \\
\hline
allenai/wildjailbreak~\citep{wildteaming2024} & 2{,}000 & 210 & 2{,}210 \\
\hline
LLM Past Tense~\citep{andriushchenko2024refusal} & 730 & 0 & 730 \\
\hline
Necent/llm-jailbreak-prompt-injection-dataset~\citep{Necent} & 629 & 371 & 1{,}000 \\
\hline
TrustAIRLab/JailbreakQR~\citep{chu2025jades} & 400 & 0 & 400 \\
\hline
\textbf{Total} & \textbf{4{,}243} & \textbf{1{,}097} & \textbf{5{,}340} \\
\hline
\end{tabular}
\end{table}

\section{Dataset Distributional Similarity}
\label{appsec:js_divergence}

\begin{figure}[h]
\centering
\includegraphics[width=0.9\linewidth]{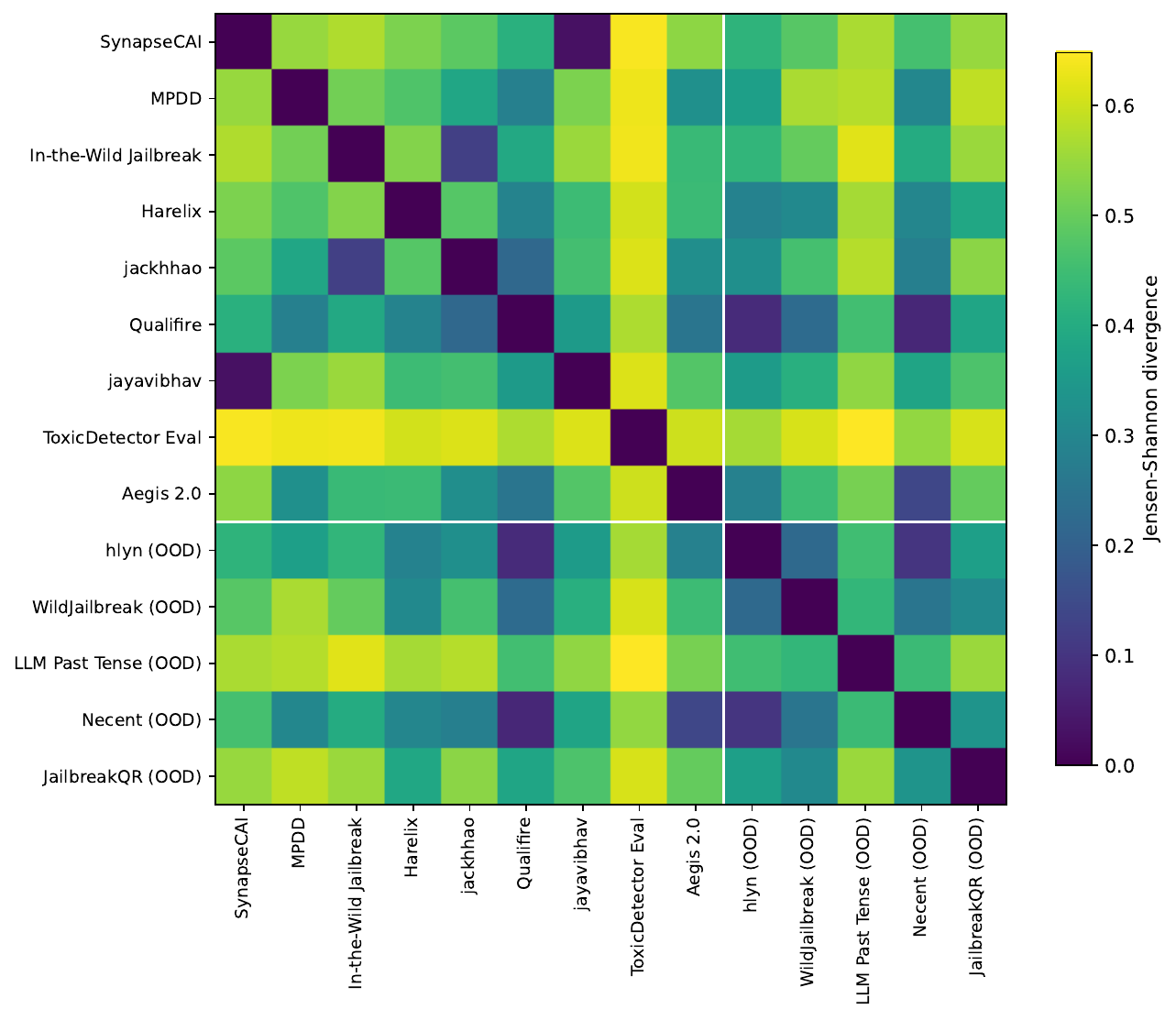}
\caption{Pairwise Jensen-Shannon divergence between the nine training datasets and the five OOD test datasets. Prompts are embedded with \texttt{all-mpnet-base-v2}, the shared embedding space is partitioned into 50 clusters via $k$-means, and each dataset's distribution is represented as its normalized histogram of prompts per cluster. Darker cells indicate greater distributional similarity between a pair of datasets.}
\label{fig:js_divergence}
\end{figure}

To provide additional evidence for the distributional separation between \sysname's training sources and the source-disjoint OOD test set, Figure~\ref{fig:js_divergence} shows pairwise Jensen-Shannon divergence between the prompt-text distributions of all nine training datasets and all five OOD test datasets. Higher divergence (yellow) indicates greater dissimilarity between two datasets' content; lower divergence (dark purple) indicates the two datasets' prompts are drawn from more similar distributions. Datasets along the diagonal trivially show zero divergence against themselves.

\texttt{necent} and \texttt{hlyn} exhibit comparatively lower divergence against several training datasets, consistent with their construction as aggregations over a much wider pool of public sources, where some degree of content overlap with any individual constituent dataset is expected.

\section{Baseline Details}
\label{appsec:baseline_details}

\begin{itemize}
    \item \textbf{OpenAIModerationAPI}~\citep{markov2023holistic}{:} A widely deployed, industrial-grade black-box API designed to detect text that violates safety policies. It comprises an unknown number of parameters although it likely has more parameters than our method due to the API's ability to further classify the exact type of content policy violation, such as `hate', `self-harm', `harassment', `violence' etc. It is otherwise unable to be tweaked rapidly to specific datasets and newer attacks.
    \item \textbf{ToxicDetector}~\citep{liu2024efficient}\textbf{:} A grey-box methodology designed to perform binary classification for benign or toxic prompts. It uses embeddings from an LLM (such as Llama-2-7B, which we use in our experiments) as the feature vectors for a 300M parameter MLP classifier.
    \item \textbf{ShieldGemma}~\citep{zeng2024shieldgemma}\textbf{:} These are white-box, instruction-tuned models for evaluating the safety of text. ShieldGemma 1 is built upon the Gemma 2 LLM in 2B, 9B, and 27B parameter sizes (we use the 2B one in our experiments), and allows for a custom safety policy to be provided in the system prompt.
    \item \textbf{LastLayer}~\citep{arekusandr_last_layer}\textbf{:} This is a partially black-boxed security library for protecting LLMs from malicious attacks. Rather than employing a deep-learning-centric approach, LastLayer holistically analyses the structure of incoming prompts via numerous modules, in order to detect patterns indicative of prompt injection attacks, jailbreaks and other exploits.
    \item \textbf{StruQ} \citep{chen2025struq}: A white-box defense against prompt injection that separates instructions from data via structured queries, using a specially instruction-tuned 7B LLM that only follows commands in the designated instruction channel and treats content in the data channel as inert. We adapt it as a detector by flagging any input whose structured completion deviates from expected instruction-following behavior. StruQ is "fine-tunable" only in the sense of retraining the full 7B-parameter LLM via structured instruction tuning, unlike \sysname's lightweight per-attack classifier fine-tuning.
\end{itemize}

\section{Discussion: Energy Efficiency and Sustainability}
\label{appsec:energy}

As AI and specifically deep learning become increasingly ubiquitous, growing concerns have emerged regarding the energy costs of deploying large models and their environmental impact~\citep{gholami2024ai, canziani2016analysis, anthony2020carbontracker, thompson2020computational}. Data center energy consumption in the U.S. accounted for 4.4\% of total electricity use and is projected to triple by 2028~\citep{shehabi20242024}. Critically, the majority of energy costs in AI systems stem from inference rather than training, as training is typically a one-time process while inference must be sustained continuously at scale~\citep{desislavov2023trends}. Despite this reality, most research has focused on reducing training costs rather than inference costs~\citep{verdecchia2023systematic}.

\sysname implicitly accounts for these sustainability concerns through its architectural design. By using an 86M parameter base safety classifier and deploying only a small subset of promptcops per inference (typically 5 out of a larger ensemble), \sysname significantly reduces both the computational footprint and energy consumption compared to billion-parameter alternatives. The framework's reliance on bootstrap aggregation and lightweight learned routing, demonstrates that powerful yet efficient methods can achieve strong performance in LLM safety without requiring massive computational resources. This approach not only makes robust LLM safety more accessible to resource-constrained deployments and smaller vendors, but also represents a step toward environmentally sustainable AI safety practices. As LLM systems continue to scale globally, such efficiency-focused architectures will become essential for balancing safety requirements with environmental responsibility.

\section{Artifact Licenses and Intended Use}
\label{app:artifacts}

Our experiments use publicly available models, datasets, and software artifacts. We use Prompt Guard 2 as the base prompt-safety classifier, and we follow the usage terms associated with the released model checkpoint. We also evaluate against publicly accessible or documented baselines, including OpenAI Moderation API, StruQ, ToxicDetector, ShieldGemma, and LastLayer, under their respective usage terms.

For the datasets used in our experiments, we rely on the licenses and terms provided by the original dataset creators or hosting platforms. Our released code is intended for research and reproducibility, and any redistribution or reuse of third-party artifacts remains governed by the original artifact licenses.

We use all third-party artifacts in a manner consistent with their intended research and evaluation purposes. The prompt-safety models and baseline detectors used in this work are applied only for malicious prompt detection and comparative safety evaluation, which aligns with their intended role as prompt moderation or guardrail systems. The datasets used in our experiments are used for research on malicious-prompt detection, robustness, and LLM safety evaluation, rather than for generating or facilitating harmful model behavior.

\end{document}